# Multilingual Scene Character Recognition System using Sparse Auto-Encoder for Efficient Local Features Representation in Bag of Features

Maroua Tounsi [a,x], Ikram Moalla [a,y], Frank Lebourgeois[z], Adel M. Alimi [a]

[a] REsearch Groups in Intelligent Machines Laboratory, ENIS-Sfax, Tunisia

[x]Higher Institute of Computer Science and Communication Techniques, University of Sousse, Tunisia

[y] Al Baha University, Saudi Arabia

[z]Laboratoire d'InfoRmatique en Images et Systmes d'information (LIRIS), INSA of Lyon, France
Emails:
*tounsi.maroua@ieee.org, ikram.moalla@ieee.org, frank.lebourgeois@liris.cnrs.fr, adel.alimi@ieee.org*

**Abstract**—The recognition of texts existing in camera-captured images has become an important issue for a great deal of research during the past few decades. This give birth to Scene Character Recognition (SCR) which is an important step in scene text recognition pipeline.

In this paper, we extended the Bag of Features (BoF)-based model using deep learning for representing features for accurate SCR of different languages. In the features coding step, a deep Sparse Auto-encoder (SAE)-based strategy was applied to enhance the representative and discriminative of image features. This deep learning architecture provides more efficient features representation and therefore a better recognition accuracy.

Our system was evaluated extensively on all the scene character datasets of five different languages. The experimental results proved the efficiency of our system for a multilingual SCR.

———————————— F ————————————

## 1 INTRODUCTION

The displayed amounts of texts in our background, such as text on road signs, shop names and product advertisement presents information for people and presents an essential tool to interact with their environment. Hence, the need to an automatic OCR in the wild has become mandatory, specifically for a growing number of vision applications, such as automatic text translator used by tourists in foreign countries, automatic sign recognition and visual-based navigation, among others.

Scene Character Recognition (SCR) is an important step in text recognition pipeline. Therefore, a special focus was put on SCR issues and a great deal of research in this field has been proposed during the past few years [1–4].

The task of SCR has been found to be more difficult and more complex than that of recognizing existing in scanned documents. While in these documents, characters are almost the same size when dealing with the same paragraph or title and do not show any stretch, rotation or contrast, scene characters are very specific and have variable sizes, forms styles or colors. They could be illuminated, stretched, distorted, with a non-planar surface and changeable back-ground. Scene characters may therefore consist on a very variable set of shapes, and thus induces a wide intra-class variability (in the case of the same character) in addition to inter-class variability (between characters).

Furthermore, in the absence of a specific context, the words to be recognized may belong to different languages and/or scripts.

This is our main underlying motivation for looking for a robust system invariant to scale, rotation, stretch, brightness, and so on, which is effective for the SCR domain in a multilingual and multi-script context.

Most recently published methods consider the scene characters as a special category of objects and therefore shared many of the techniques used to recognize objects. Furthermore, BoF framework seems to be an efficient model for object recognition. So, the idea was to take advantage of the feature extraction and representation methods that perform well in object recognition tasks to achieve the SCR. However, a major problem hinders the performance of such a pipeline. In fact, the information loss may result in a loss of discrimination between classes and therefore a lack of spatial information.

To deal with these problems, in this paper we proposed a novel deep feature learning architecture based on Sparse Auto-Encoder (SAE) [5]. We tried to benefit from the strengths of the BoF framework and the power of the SAE in features representation. This technique provides a more

  

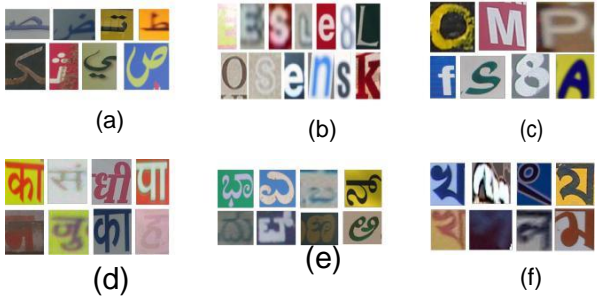

(a)   (b)   (c)

(d)   (e)   (f)

Fig. 1: Samples of scene characters of different languages: (a) English scene characters from Chars74K (b) English scene characters from ICDAR2003, (c) Arabic scene characters from ARASTI, (d) Devanagari scene characters from DSIW-3K, (e) Kannada scene characters from Chars74K, and (f) Bengali scene characters from images captured in West Bengal, India.

concise visual dictionary, more efficient features learning and a better recognition accuracy, compared to the other features representation techniques like the sparse coding. Specifically, we constructed a two-hidden-layer using the SAE in order to fine-tune the visual dictionary and learn the discriminative feature codes for each class.

The contribution of this paper is threefold. First, we handled five language scene character recognition problems, namely, Arabic, English, Bengali, Devanagari and Kannada. Indeed, we used a patch-based method which allowed us to be script independent and enabled us to deal with different language scene characters. Second, we applied stacked SAE for sparse and discriminative features representation in a BoF framework. Third, we provided a suitable benchmark to compare the performances of different methods for Arabic scene text reading.

The rest of the paper is organized as follows. Section 2 introduces our methodology. Section 3 presents a description of our system. The experimental results, datasets and discussion are then presented in section 4. Our concluding remarks and future work are finally presented in section 5.

## 2 METHODOLOGY

In this paper, we use a hybrid architecture that merges the complementary strength of the BoF framework and deep architectures was used. Furthermore, a patch-based method was applied allowing us to be script independent when dealing with different language scene characters. In this section, we introduced our work and we reviewed the related techniques.

### 2.1 Scene character recognition based on BoF

It is worth noting that the BoF framework seems to be an efficient model for object recognition.

Meanwhile, in the BoF framework, the major problem is the information loss brought by feature coding leading to the loss of discrimination between classes and the lack of spatial information. This is because many patches of a character may appear in another location of another character. Ignoring spatial information can produce character classification confusion. For example, for English characters, 'B' can be overlapped with 'R' (figure 2) if one ignores the context above this shared segment.

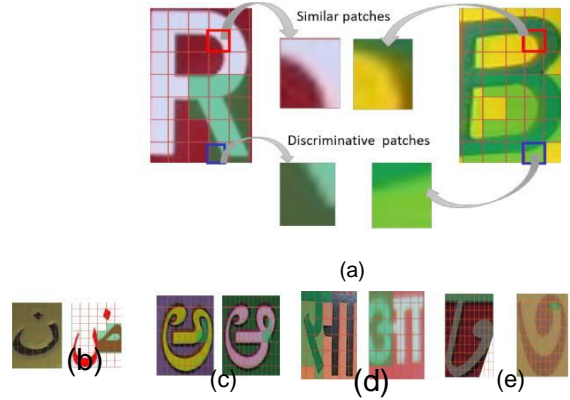

Fig. 2: Some examples of characters with discriminative patches (in green) which have the same context and can be the origin of confusion without keeping spatial information. Examples of confusion in (a) English characters, (b) Arabic characters, (c) Kannada characters, (d) Devanagari characters and (e) Bengali characters.

Therefore, character models of the same character in different positions need to be discriminant from each other. So, it is desirable to give priority to the patches containing more discriminative features and ignore those with low discriminative power in the global classification.

How to alleviate the above problems in the BoF-based framework to enhance the representative and discriminative abilities of image features while incorporating the spatial information for SCR problem is a challenging and also becomes our focus in this paper.

### 2.2 Scene character recognition based on deep learning

So far, learning features directly from data through deep learning methods has become increasingly popular and effective for visual recognition.

When applied to character recognition, the deep learning exploits the unknown structure of a character in the input distribution in order to discover good representations, often at multiple levels, with higher-level learned features defined in terms of lower-level features.

To deal with the information loss problem, deep architectures have recently proven recently their effectiveness in reducing the information loss by integrating unsupervised pre-training [6].

### 2.3 Scene character recognition based on stroke based models

Certain characters or parts of characters have low information value for recognition used in the text as they appear in multiple characters (see figure 2), whereas other characters or their parts are more discriminative by being unique to only one or a small number of scripts. Therefore, it is desirable to give priority to the patches containing more discriminative features and ignore those with a low discriminative power in the global classification. Furthermore, to deal with the multilingual SCR we used a patch-based method which allows it to be script independent dealing with different language scene characters and also to be able to treat the loss of discrimination between classes and the lack of spatial information.

To deal with this problem, many works have exploited the stroke based models for scene characters in a diversity of languages and has demonstrated competitive results. They tried to improve previous results by responding to the need of keeping the spatial information in recognizing characters and discovering the important spatial layout of each character class from the training data in a discriminative way.

In this context, several related works which deal with the SCR problem using the BoF or deep learning or stoke based models can be presented. DeCampos et al. [7] proposed to apply a BoF-based model on scene characters and compare the effectiveness of different local features. Zheng et al. [8] proposed a local description-based method to measure the similarity between two character images and use the Nearest Neighbor-based method to build a system for Chinese character recognition. In [9], the authors recognized Devanagari characters by comparing 6 types of local feature descriptors namely, pixel density, directional features, HOG, GIST, Shape Context and Dense SIFT.

While many works exploited the BoF in recognizing scene characters in a diversity of languages and demonstrated competitive results, applying this framework can admit an extension to improve the results and respond to the need of keeping spatial information in recognizing characters.

Different network structures have been studied in [10]–[12] to obtain a better recognition performance by reducing the information loss through varying the number of layers, the activation function and the sampling techniques.

Although the above-mentioned works demonstrate competitive results, the used methods require millions of annotated training samples to perform well. However, these annotated training images are not publicly available.

Our work is also related to the stroke based models where regions which contain the basic information are learned. We quote as an example the work of Shi et al. [13] wich extend [14] to a discriminative multiscale stroke detector-based representation (DMSDR). Tian et al. [3] learn the spatial information by using the cooccurrence of a histogram of oriented gradients (Co-HOG) which captures the spatial distribution of neighboring orientation pairs. Chen-Yu Lee et al. [15] propose a discriminative feature pooling method that automatically learns the most informative strokes of each scene character within a multi-class classification framework, whereas each sub-region seamlessly integrates a set of low-level image features through integral images.

While many works used the BoF or the deep learning based method or stroke-based model for SCR, there is not any single attempt that exploits the deep learning for feature representation in a BOF for SCR using a patch-based model.

# 3 LEARNING FEATURE REPRESENTATION IN A BOF

In this section, a detailed description of our architecture based on the use of an adopted BoF model using SAE to encode local descriptors was introduced.

In this paper, we used a hybrid architecture that merges the complementary strength of the BoF framework and deep architectures. Figure 3 shows our BoF framework using a SAE to learn visual codes and perform the feature coding. During training, the SAE was first used to perform an unsupervised learning of the local features. We concurrently learnt a local classifier, while back-propagating the residual error to fine-tune the codebook. After training, the SAE was used as a feed-forward encoder to directly infer each local feature coding.

## 3.1 SAE for unsupervised features representation in BoF framework

After extracting local patches and learning local features via local descriptors recognized by their invariance and their compact representation of images, we perform the unsupervised features representation. In fact, the number of nodes for both the input and output layers is d, which is the same as the descriptor dimension. In the BoF, we have K codewords in the codebook W, therefore, we also set the number of hidden nodes in the AE network as K, which means that the coded vectors of AE network and other BoF models have the same discrete dimensionality (K can be set as 1024, 2048, etc.). Let $V \in R^{d*K}$, $b(1) \in R^K$ be the weight and bias of the input-hidden layer.

### 3.1.1 Constructing a single layer

An auto-encoder network [16] is a variant of a three-layer MLP, which models a set of inputs through unsupervised learning. The auto-encoder learns a latent representation that performs reconstruction of the input vectors. This means that the input and output layers assume the same conceptual meaning. Given an input x, a set of weights W maps the inputs to the latent layer z. From this latent layer, the inputs ($\hat{x}$) are reconstructed with another set of weights V. The feature layer has a dimension d corresponding to the size of the input vector (e.g. 128 for SIFT). The coding layer has K visual codewords used to encode the feature layer. The layers are connected via undirected weights W, our visual dictionary. Given a visual dictionary W, feature x can be encoded to its visual code z. The learning is concurrently regularized to guide the representation to be sparse. Specifically, suppose we have $W \in R^{d*N}$, $b(1) \in R^N$ is the weight and bias of the input-hidden layer, $V \in R^{N*d}$, $b(2) \in R^d$ is the weight and bias of the hidden-output layer. Suppose we have extracted N descriptors: $X = [x_1, x_2, \cdots, x_N] \in R^{d*N}$ from the training images around N interesting points by dense sampling. $N = \sum_{i=1}^{N} x_i$, denotes the N input data where K is the number of neurons in the hidden layer which corresponds to the size of visual dictionary W, and $\hat{x}_i \in R^{w*K}$ denotes the reconstruction of $x_i$, our objective is to learn a shared mid-level visual dictionary. $W = [w_1, w_2, \cdots, w_K] \in R^{w*K}$ and a reconstruction weight $V = [v_1, v_2, \cdots, v_K] \in R^{v*K}$ to make $x_i$ to be as close to $x_i$

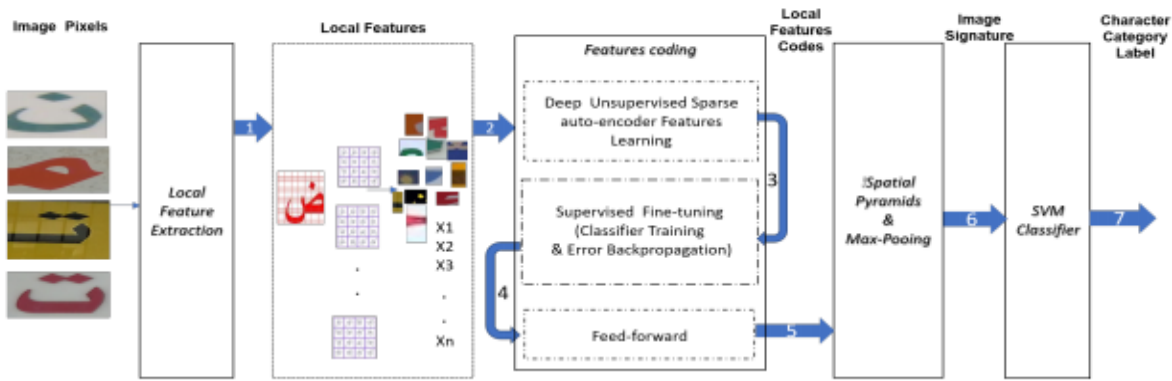

Fig. 3: Our SAE-based BoF architecture for SCR. The arrows indicate the operations performed during the various phases.

as possible, by minimising the following objective function $E_1(W, V; X)$

$$E_1(W, V; X) = \frac{1}{2N} \sum_{i=1}^{N} \|x_i - \hat{x}_i\|_2^2 + \beta \sum_{k=1}^{K} KL(\rho \|_{\hat{\rho}_k}) \quad (1)$$

$$\hat{x}_i = V(1 + exp(W^T x_i))^{-1} \quad (2)$$

$$KL(\rho \|_{\hat{\rho}_k}) = \rho \log \frac{\rho}{\hat{\rho}_k} + (1 - \rho) \log \frac{1 - \rho}{1 - \hat{\rho}_k} \quad (3)$$

$$\hat{\rho}_k = \frac{1}{N} \sum_{i=1}^{N} (1 + exp(w_k^T x_i))^{-1} \quad (4)$$

Where $\|.\|_2$ indicates the L2-norm, $\rho$ is a sparsity parameter, W is the mid-level visual dictionary to be learned which is subject to $\|w_k\|^2 = 1, \forall k = 1, 2, \cdots, K$, to prevent V from being arbitrarily large, V is a reconstruction weight which reconstructs the input layer from the hidden layer, $\beta$ is a parameter controlling the weight of the sparsity penalty term, $(\hat{\rho}_k)$ is the average activation of the $k^{th}$ hidden node averaged over the N training data and $\rho$ is the target average activation of the hidden nodes. The KullbackLeibler divergence KL(.) is a standard function for measuring how different two distributions are. It has the property that $KL(\rho \|_{\hat{\rho}_k}) = 0$, if $(\hat{\rho}_k) = \rho$, otherwise it increases monotonically as $(\hat{\rho}_k)$ diverges from $\rho$. Hence, the KullbackLeibler divergence KL(.) is used for the sparsity constraint.

### 3.1.2 Constructing a deep architecture

We extend the single-layer architecture to be hierarchical. In the case of deep fully-connected networks, we stack two SAE. We greedily stack one additional auto-encoder with a new set of parameters V that aggregates within a neighborhood of outputs from the first auto-encoder W. This allows the representation to be transformed from a low-dimensional to a single high-dimensional vector that describes the entire image.

### 3.2 Spatial pyramid matching and max pooling

Using the SPM technique, each image was segmented into fine sub-regions. Then, we calculated the histogram of the local features of each sub-region. We segmented the images using different scales and computed the BoF histogram in each segment. After that, we concatenated all the histograms. Using this strategy, the performance was considerably improved over the basic bag-of-features representation. We applied then a max pooling technique in order to summarize all the local features representing the image. We concatenated the pooling features in different levels of the pyramid and the final image level features or representations were built.

### 3.3 Scene character recognition

Scene character images segmented from a text in the wild can be recognized by training a suitable classifier. The linear Support Vector Machine (SVM) seems to be the most efficient machine learning technique used for high dimensional data since it is fast and capable of providing the state of-the-art recognition rates. Therefore, in this work, we have chosen to use the linear SVM with L2-regularization to classify signatures into their categories.

## 4 EXPERIMENTAL RESULTS

### 4.1 Experimental setting

The images were first resized to 90*90 pixels. SIFT descriptor (Scale Invariant Feature Transform) [17] was densely sampled from each image. We randomly sampled 200,000 patches for ICDAR03-CH and Chars74K English datasets; 80,000 patches for ARASTI dataset; and 150,000 patches for Chars74K Kannada, Bengali and Devanagari datasets. The trained dictionary was used to encode the local features.

### 4.2 Multilingual Scene Character Datasets

This section described the several scene characters used in the evaluation of this work. These are six scene character datasets of five different languages including two sets in English, one in Arabic, one in Devanagari, one in Kannada and one in Bengali.

### 4.2.1 ARASTI: An Arabic scene character dataset

In spite of the diversity of the works interested in text recognition in natural scenes for a diversity of languages there is a scarcity in publicly available OCR datasets for scene text recognition. Thus, the creation of scene character datasets of Arabic would have a significant impact on the advancement of scene text recognition research. Taking this novelty into account, we have initiated the development of a real dataset of annotated images of text in diverse natural scenes captured at varying weather, lighting and perspective conditions for Arabic scene text recognition, called ARASTI (Arabic Scene Text Image) dataset [18]. To our knowledge, ARASTI is the first dataset for Arabic scene text recognition in real-world. Our ARASTI dataset is freely available[2].

### 4.2.2 English scene characters datasets

We evaluated the proposed recognition system on two English scene character datasets, i.e, Chars74K [7] [3] and ICDAR03-CH [19] datasets.

In order to be able to compare our results to previously published findings we concentrated on testing the Chars74K benchmark 15 training images per class, and the ICDAR03-CH benchmark with 15 training images per class.

### 4.3 Hindi scene characters datasets

Bengali scene characters dataset

The Bengali alphabet set has 50 basic characters which include 11 vowels and 39 consonants. Additionally, it has several diacritics and a large number of conjunct characters. In this paper, we used the dataset proposed in [3]. Some of these Bengali character samples are shown in figure 1f.

Devanagari scene characters dataset In this paper, we used the DSIW-3K [9] to evaluate the scene Devanagari recogni-tion rate (see figure 1d).

Kannada scene characters dataset Kannada is an Indic script language. We used Kannada Chars74K dataset [4] [20] for our experiments. (See figure 1e).

### 4.3.1 Evaluations of each part in the system and the parameters

### 4.3.2 Evaluation of the dictionary size

Figure 4 shows that a larger codebook has more capacity to capture the diversity in the features, but is also more likely to exhibit codeword redundancies. From our experiments, 1024 codewords seem to give a good balance between diversity and conciseness ().

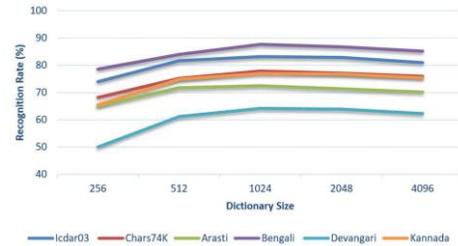

Fig. 4: Recognition rate when varying the dictionary size.

### 4.3.3 Evaluation of the SAE learning strategy: unsupervised versus supervised strategies

In this section, we discussed the empirical performance of both depth and supervision, and offered possible explanations of the results. We began with the basic shallow unsupervised architecture (figure 5), which allowed us to achieve close to the state-of-the-art results. We evaluated the unsupervised and supervised SAE learning strategy on ARASTI and ICDAR03-CH datasets. We compared the results of supervised and unsupervised learning with similar dictionary size. The deep supervised architecture produced the best recognition accuracy.

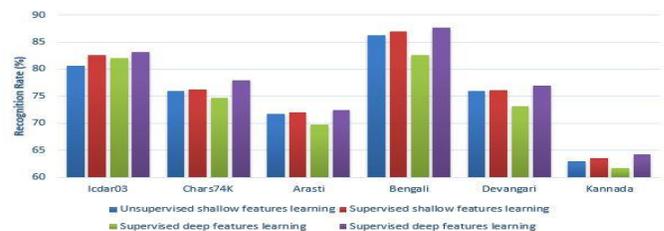

Fig. 5: Results for the shallow and deep architecture with and without supervised fine-tuning (%).

### 4.4 Comparaison with previously published results using CNN based methods

Table 1 presents the character recognition accuracy on the two public English scene character datasets. It shows that our proposed methods greatly outperform the other previously published results using CNN based methods. The results show that ConveNet [21] achieves an accuracy of 86.5%, whereas our system could reach 88.5% with fewer of training samples.

|  | ICDAR03-CH | Chars74K |
|---|---|---|
| LTP+MSER+CNN [21] | 86.5 | - |
| unsupervised features + CNN [22] | 83.9 | - |
| CNN [12] | 79.5 | 75.3 |
| **Proposed method** | **88.5** | **85.4** |

TABLE 1: The performance of our method compared to previously published results using CNN based methods.

### 4.5 Comparative study with state-of-the-art methods

### 4.5.1 Results on English scene character datasets

Table 2 shows the results obtained with training sets of 15 samples per class and test sets of 5 samples per class. Table 2 reports on our proposed system and some existing systems. It can be concluded that our proposed system using Stacked Sparse Auto-encoder (SSAE) for supervised features learning gives better results than those of the existing methods. For further details, the confusion matrix is displayed in figure 6 where we notice that upper case letters and lower case letters like 'X' and 'x', 'W' and 'w', 'O' and 'o', 'Z' and 'z' were confused. Meanwhile, these characters are extremely difficult to distinguish, even by human eyes.

TABLE 2: The performance of our system compared to previously published results using other methods on English script (%). Stacked Sparse Auto-encoder (SSAE).

| | ICDAR03-CH | Chars74K |
|---|---|---|
| Convolutional Co-occurrence of HOG [3] | 81.7 | - |
| CoHOG+Linear SVM [23] | 79.4 | - |
| PHOG+Linear SVM [24] | 76.5 | - |
| GHOG + SVM [25] | 76 | 62 |
| LHOG + SVM [25] | 75 | 58 |
| HOG + NN [26] | 52 | 58 |
| SIFT+ SVM [7] | - | 21.40 |
| Strokelets [27] | 69 | 62 |
| Co-occurrence of strokelets [28] | 82.7 | 67.5 |
| PCAnet[29] | 75 | 64 |
| HOG + Sparse Coding + SVM[30] | 72.27 | - |
| SIFT + Sparse Coding+ SVM [31] | 75.3 | 73.1 |
| SIFT + SAE + SVM [32] | 79.3 | 75.4 |
| SIFT + Sparse RBM + SVM [32] | 78.1 | 73.9 |
| **SIFT + SSAE + SVM** | **82.2** | **76.6** |
| **SIFT + SSSAE + SVM** | **83.2** | **77.9** |

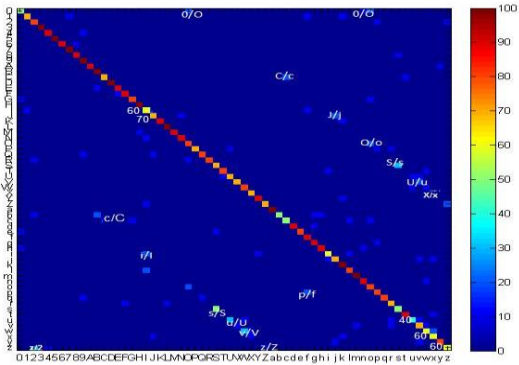

Fig. 6: Confusion matrix resulting from ICDAR03-CH dataset.

### 4.5.2 Results on ARASTI dataset

This subsection detailed the recognition result of Arabic characters. In our experiments, we compared the results obtained using a conventional OCR system; Tesseract OCR[5]. Arabic scene character recognition is a very challenging task because some characters are distinguished from each other by the number and the position of dots around the basic shape of the character. To alleviate this problem, the similar classes were merged into one. So, the samples from the 55 initial classes (denoted by 'ARASTI-55' in table 3) with similar structures were combined. Hence, a total of 30 classes (denoted by 'ARASTI-30' in table 3) to be recognized were obtained; they are shown in the confusion matrix in

5. https://code.google.com/p/tesseract-ocr/

figure 7. Table 3 shows the results obtained with training sets of 15 samples per class and test sets of 5 samples per class.

We can conclude that our system yields competitive results and prove that SSAE is a good choice for Arabic character recognition. Its good performance could be explained by the fact that the SSAE is designed to learn the discrim-inative patches and keep the spatial information without information loss. For further details, figure 7 introduces the confusion matrix.

TABLE 3: The performance of our system on Arabic scene characters dataset compared to previously published results using other methods (%). Stacked Sparse Auto-encoder (SSAE).

| | Arasti-55 | Arasti-30 |
|---|---|---|
| Tesseract OCR | 19.6 | 22.3 |
| Tounsi et al. [31] | 57.5 | 60.4 |
| Tounsi et al. [32] | 61.2 | 65.1 |
| **SIFT + SSAE + SVM** | 61.9 | 71.3 |
| **SIFT + SSSAE + SVM** | **62.9** | **72.5** |

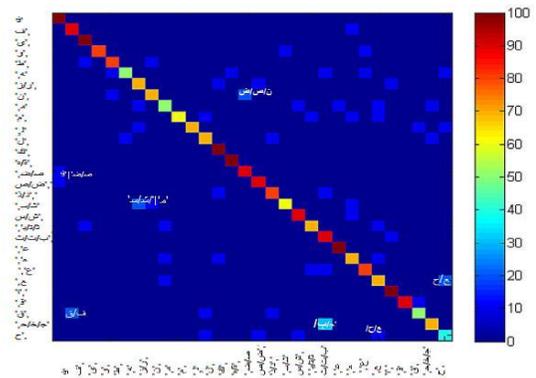

Fig. 7: Confusion matrix resulting from ARASTI dataset

### 4.5.3 Results on Hindi script scene character datasets.

TABLE 4: The performance of our system in recognition on Bangali, Kannada and Davangari scene characters datasets compared to previously published results using other methods (%). Stacked Sparse Auto-encoder (SSAE).

| | Bengali | Kannada | Devanagari |
|---|---|---|---|
| Jaderberg et al. (CNN) [12] | 84.5 | 76.2 | 62.7 |
| Sheshadri et al. [20] | - | 54.13 | - |
| deCampos et al. [7] | - | 29.88 | - |
| Narang et al. [9] | - | - | 56.0 |
| Tounsi et al. [31] | 84.2 | 72.5 | 48.4 |
| Tounsi et al. [32] | 87.7 | 76.5 | 58.7 |
| **SIFT + SSAE + SVM** | 85.2 | 76.5 | 64.0 |
| **SIFT + SSSAE + SVM** | **89.2** | **77.1** | **64.2** |

Table 4 lists the obtained results on Hindi scripts datasets. The Tesseract OCR result is not available since it cannot deal with isolated Hindi characters. As shown in table 4, the recognition accuracy outperforms the current state-of-the-art on Bangali, Kannada and Devanagri datasets even with CNN [12]. This is due to the training data which is insufficient which makes the CNN suffer from over fit-ting. Our proposed system demonstrates its efficiency even with insufficient training data and with different languages. Some successful and failed cases are shown in figure 8.

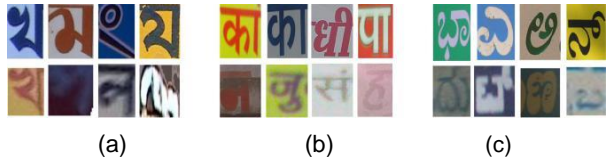

(a)                    (b)                    (c)

Fig. 8: Successful and failed cases of our system. In each figure, the top row shows the correctly recognized samples and the bottom row shows wrongly recognized ones. (a) Bengali language, (b) Devanagari language and (c) Kannada Language.

## 5 CONCLUSION

In this paper, a deep learning strategy based system in BoF was proposed in order to fine-tune the visual dictionary and to learn local features codes. This provides more effi-cient features representation and thus a more recognition accuracy, compared to the other feature representation tech-niques such as the sparse coding.

The proposed system enabled us to tackle an important problem resulting from text recognition in images of natural scenes: Multilingual SCR. In fact, we handled five language scene character recognition problems, namely, Arabic, English, Bengali, Devanagari and Kannada towards the advancement of scene character recognition research for multiple scripts. In addition, we provided a benchmark results for comparing Arabic scene characters recognition methods. Our future research will mainly include: first, our frame-work will be adapted to word-level text recognition and also we will further investigate recognizing scene text in the wild in natural video scenes.

### ACKNOWLEDGMENT

This work is performed in the framework of a thesis MO-BIDOC financed by the EU under the program PASRI. The authors would like also to acknowledge the partial financial support of this work by grants from General Direction of Scientific Research (DGRT), Tunisia, under the ARUB program. The research leading to these results has received funding from the Ministry of Higher Education and Scien-tific Research of Tunisia under the grant agreement number LR11ES48.